\documentclass{article}

\usepackage[final]{corl_2020}

\usepackage[utf8]{inputenc} % allow utf-8 input
\usepackage[T1]{fontenc}    % use 8-bit T1 fonts
\usepackage{hyperref}       % hyperlinks
\usepackage{url}            % simple URL typesetting
\usepackage{booktabs}       % professional-quality tables
\usepackage{amsfonts}       % blackboard math symbols
\usepackage{nicefrac}       % compact symbols for 1/2, etc.
\usepackage{microtype}      % microtypography
\usepackage{graphicx}
\usepackage{xcolor}
\usepackage{xspace}
\usepackage{wrapfig}
\usepackage{amsmath}
\usepackage{multirow}
\usepackage[font=normalsize]{caption}
\usepackage{tabularx}

\makeatletter
\DeclareRobustCommand\onedot{\futurelet\@let@token\@onedot}
\def\@onedot{\ifx\@let@token.\else.\null\fi\xspace}

\newcommand{\model}{TNT\xspace}
\newcommand{\bfs}{\textbf{s}}
\newcommand{\bfx}{\textbf{x}}
\newcommand{\bfc}{\textbf{c}}

\def\eg{\emph{e.g}\onedot}

\def\etc{\emph{etc}\onedot} 
 
\def\etal{\emph{et al}\onedot}
\makeatother

\title{TNT: Target-driveN Trajectory Prediction}

\newcommand*\samethanks[1][\value{footnote}]{\footnotemark[#1]}

\author{%
  Hang Zhao$^{1}$ \thanks{Equal contribution. Correspond to \texttt{\{hangz, jiyanggao\}@waymo.com}.} \qquad Jiyang Gao$^{1}$ \samethanks \qquad Tian Lan$^{1}$ \qquad Chen Sun$^{2}$ \qquad Benjamin Sapp $^{1}$ \\
  \textbf{Balakrishnan Varadarajan$^{1}$ \qquad Yue Shen$^{1}$ \qquad Yi Shen$^{1}$ \qquad Yuning Chai$^{1}$ } \\ 
  \textbf{Cordelia Schmid$^{2}$ \qquad Congcong Li$^{1}$ \qquad Dragomir Anguelov$^{1}$} \\ \\
  $^1$Waymo LLC ~~ $^2$Google Research
}

\begin{document}

\maketitle

\vspace{-1em}
% \begin{center}
%  $^1$Waymo LLC ~~ $^2$Google Research
% \end{center}

\begin{abstract}
Predicting the future behavior of moving agents is essential for real world applications. It is challenging as the intent of the agent and the corresponding behavior is unknown and intrinsically multimodal. Our key insight is that for prediction within a moderate time horizon, the future modes can be effectively captured by a set of target states. 
This leads to our target-driven trajectory prediction (\model) framework. 
\model has three stages which are trained end-to-end. It
first predicts an agent's potential target states $T$ steps into the future, by encoding its interactions with the environment and the other agents. 
\model then generates trajectory state sequences conditioned on targets. A final stage estimates trajectory likelihoods and a final compact set of trajectory predictions is selected.
This is in contrast to previous work which models agent intents as latent variables, and relies on test-time sampling to generate diverse trajectories.
We benchmark \model on trajectory prediction of vehicles and pedestrians, where we outperform state-of-the-art on Argoverse Forecasting, INTERACTION, Stanford Drone and an in-house Pedestrian-at-Intersection dataset.
\end{abstract}

\keywords{Trajectory prediction, multimodal prediction.} 
\section{Introduction}
% Raise the problem.
Predicting the future states of moving agents in a real-world environment is an important and fundamental problem in robotics.
For example, in the setting of autonomous driving on public roads, it is essential to have an accurate understanding of where the other vehicles and pedestrians will likely be in the future, in order for an autonomous vehicle to take safe and effective actions.

% Talk about the challenges.
A key challenge to future prediction is the high degree of uncertainty, in large part due to not knowing the intents and latent characteristics of the other agents. For example, a vehicle commonly has a multimodal distribution of futures: it could turn, go straight, slow down, speed up, \etc. Depending on other scene elements, it could pass, yield, change lanes, or pull into a driveway.
% Overview of previous solutions and weaknesses.
This challenge has garnered a lot of interest in the past few years.  One approach to model the high degree of multimodality is to employ flexible implicit distributions from which samples can be drawn---conditional variational autoencoders (CVAEs)~\cite{DESIRE}, generative adversarial networks (GANs)~\cite{SocialGAN}, and single-step policy roll-out methods~\cite{precog_Rhinehart_2019_ICCV}.
Despite their competitive performance, the use of latent variables to model intents prohibits them to be interpreted, and often requires test-time sampling to evaluate probabilistic queries (\eg, ``how likely is the agent to turn left?'').
Furthermore, considerable effort has gone into addressing mode collapse in such models, in the machine learning community at large~\cite{salimans2016improvedGANS} and specifically for self-driving cars~\cite{rhinehart2018r2p2,kitani_diverse_forecasting_dpps}.

\begin{figure}[h]
    \centering
    \includegraphics[width=0.9\linewidth]{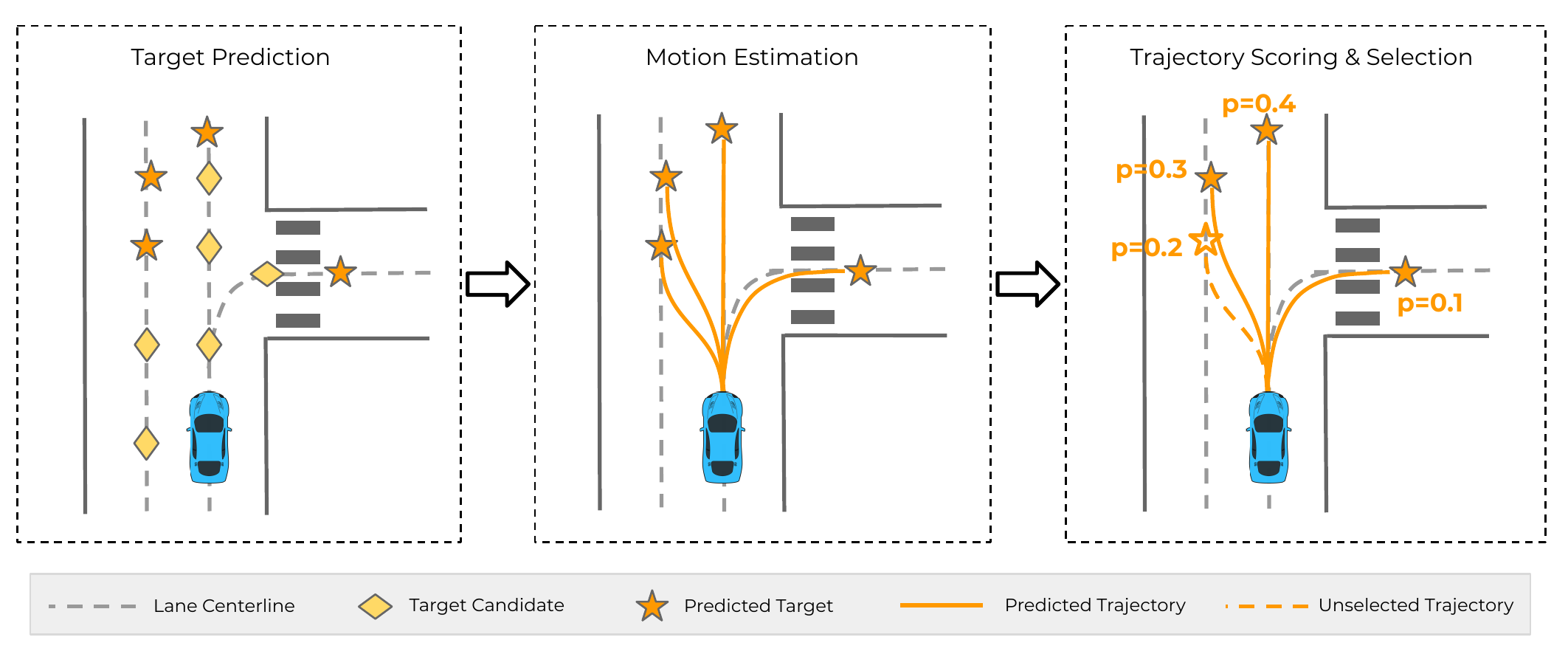}
    \caption{Illustration of the \model framework when applied to the vehicle future trajectory prediction task. \model consists of three stages: (a) {\em target prediction} which proposes a set of plausible targets (stars) among all candidates (diamonds). (b) {\em target-conditioned motion estimation} which estimates a trajectory (distribution) towards each selected target, (c) {\em  scoring and selection} which ranks trajectory hypotheses and selects a final set of trajectory predictions with likelihood scores.}
    \label{fig:teaser}
    % \vspace{-0.1in}
\end{figure}

To address these limitations, we make the observation that for our task (\eg vehicle and pedestrian trajectory prediction), the uncertainties over a moderately long future can be mostly captured by the prediction of possible \textit{targets} of the agents. These targets are not only grounded in physical entities that are interpretable (\eg location), but also correlate well with intent (\eg a lane change or a right turn). We conjecture that the space of \textit{targets} can be discretized in a scene --- allowing a deterministic model to generate diverse targets in parallel --- and later refined to be more accurate.

These observations lead to our proposed \textit{target-driven trajectory prediction} framework, named \model. We first cast the future prediction problem into predicting a distribution over discretized {\em target} states, and then formulate a probabilistic model in which trajectory estimation and likelihood are conditioned on such targets. The resulting framework has three stages that are trained end-to-end: 
(1) {\em target prediction} estimates a distribution over candidate targets given scene context; 
(2) {\em target-conditioned motion estimation} predicts trajectory state sequences per target; and 
(3) {\em scoring and selection} estimates the likelihood of each predicted trajectory, taking into account the context of all other predicted trajectories. We obtain a final set of compact diverse predictions by ranking the likelihoods and suppressing redundant trajectories.
An illustration of our three-stage model when applied to vehicle trajectory prediction is shown in Figure~\ref{fig:teaser}. Although our model is end-to-end trained, its three stage layout, with interpretable outputs at each stage,  closely follows the typical processing steps in traditional robotics motion forecasting and planning systems~\cite{tsubouchi1994behavior,broadhurst2004prediction}, thus making it easy to incorporate domain knowledge during deployment.

We demonstrate the effectiveness of \model on multiple challenging trajectory prediction benchmarks. In the driving domain we evaluate on the Argoverse Forecasting dataset~\cite{chang2019argoverse} and INTERACTION dataset~\cite{interactiondataset}; for pedestrians, the Stanford Drone dataset~\cite{stanford_drone_dataset} and an in-house Pedestrian-at-Intersection dataset. We achieved the state-of-the-art performance on all benchmarks.

\section{Related Work}

%\noindent\textbf{Trajectory Prediction.} 
Trajectory prediction has received much attention recently, especially in autonomous driving~\cite{chang2019argoverse,interactiondataset,highDdataset,NGSIM,fang2020tpnet}, in social interaction prediction~\cite{Kitani2017Activity,SocialLSTM,helbing1995social,ma2017forecasting} and sports~\cite{zheng2016generating,ZhanBasketball}. One of the key challenges is to model multimodal futures distributions. A popular approach is to model the future modes implicitly as latent variables~\cite{DESIRE,precog_Rhinehart_2019_ICCV,SocialGAN,cui2019multimodal,hong2019rules,Yeh_2019_CVPR,sun2019stochastic,tang_multifuture}, which aims at capturing the underlying intents of the agents. For example, DESIRE~\cite{DESIRE} used a conditional VAE~\cite{kingma2013auto} while PRECOG~\cite{precog_Rhinehart_2019_ICCV} used flow-based generative models~\cite{rezende2015variational}; SocialGAN~\cite{SocialGAN} proposed an adversarial discriminator to predict realistic futures; Hong \etal~\cite{hong2019rules} modeled the motion patterns with a latent Gaussian mixture model. However, the use of non-interpretable, latent variables makes it challenging to incorporate expert knowledge into the system. Furthermore, these models require stochastic sampling from the latent space to obtain implicit distributions at run time.  These properties make them less suitable for practical deployment.

Alternatively, some approaches attempted to decompose the trajectory prediction task into subtasks, with the hope that each subtask is more manageable to solve and provides interpretable intermediate results. For example, Ziebart \etal~\cite{ziebart2009planning} proposed planning-based prediction for pedestrians, they first estimated a Bayesian posterior distribution of destinations, and then used inverse reinforcement learning (IRL) to plan the trajectories. Rehder \etal~\cite{rehder2015goal} introduced the notion of \textit{goals} which are defined as short-term destinations, and decomposed the problem into goal distribution estimation and goal-directed planning. The goals were defined as mixture of Gaussian latent variables. Their followup work~\cite{rehder2018pedestrian} then demonstrated that the whole framework can be jointly trained via IRL.
Concurrent to our work, Mangalam \etal~\cite{mangalam2020not} proposed to generate \textit{endpoints} to guide the full trajectory generation. Unlike \model, their method still relies on latent variables in CVAE to model the underlying modes of the \textit{endpoints}.

Most related to \model are works that discretize the output space as \textit{intents}~\cite{casas2018intentnet} or with \textit{anchors}~\cite{chai2019multipath,phan2019covernet}. IntentNet~\cite{casas2018intentnet} manually defined several common motion categories for self-driving vehicles, such as left turn and lane changes, and learned a separate motion predictor for each intent. This manual categorization is task and dataset dependent, and may be too coarse to capture intra-category multimodality. More recently, MultiPath~\cite{chai2019multipath} and CoverNet~\cite{phan2019covernet} chose to quantize the trajectories into \textit{anchors}, where the trajectory prediction task is reformulated into anchor selection and offset regression. The anchors are either pre-clustered into a fixed set \textit{a priori}~\cite{chai2019multipath} or obtained dynamically based on kinematic heuristics~\cite{phan2019covernet}. Unlike anchor trajectories, the targets in \model are much lower dimensional and can be easily discretized via uniform sampling or based on expert knowledge (\eg HD maps). Hence, they can be estimated more reliably.
Despite their simplicity, we demonstrate that the targets are informative enough to capture most of the uncertainty in predicting future state, and our target-driven framework outperforms the anchor-based methods.
\section{Formulation}
Given a sequence of observed states for a single agent $\bfs_P=[{s}_{-T'+1},{s}_{-T'+2},...,s_0]$, our goal is to predict its future states $\bfs_F = [ s_1,s_2,...,s_T]$ up to some fixed time step $T$. Naturally, the agent interacts with an environment consisting of other agents and scene elements for context: $\bfc_P = [{c}_{-T'+1},{c}_{-T'+2},...,c_0]$. We denote $\bfx = (\bfs_P, \bfc_P)$ for brevity, thus the overall probabilistic distribution we want to capture is $p(\bfs_F|\bfx)$.

In practice, $p(\bfs_F|\bfx)$ can be highly multimodal.
For example, a vehicle approaching an intersection could turn left, go straight or change lanes.
% a pedestrian could walk tardily, hastily or stand still. 
Intuitively, the uncertainty of future states can be decomposed into two parts: the \textit{target or intent uncertainty}, such as the decision between turning left and right; and the \textit{control uncertainty}, such as the fine-grained motion required to perform a turn.
We can therefore decompose the probabilistic distribution accordingly by conditioning on targets and then marginalizing over them:
\begin{equation}
    p(\bfs_F | \bfx) = \int_{\tau \in \mathcal{T}(\bfc_P)} p(\tau | \bfx) p(\bfs_F | \tau, \bfx) d \tau,
\end{equation}
where $\mathcal{T}(\bfc_P)$ represents the space of plausible targets depending on the observed context $\bfc_P$.

Under this formulation, our main insight is that, for applications such as trajectory prediction, by properly designing the target space $\mathcal{T}(\bfc_P)$ (\eg target locations), the target distribution $p(\tau | \bfx)$ can well capture the intent uncertainty. Once the target is decided, we further demonstrate that the control uncertainty (\eg trajectories) can be reliably modeled by simple, unimodal distributions.
We approximate the target space $\mathcal{T}(\bfc_P)$ by a set of discrete locations, turning the estimation of $p(\tau | \bfx)$ primarily into a classification task. Compared with latent variational models, our model offers better interpretability in the form of explicit target distributions, and can naturally incorporate expert knowledge (such as road topology), when designing the target space $\mathcal{T}(\bfc_P)$.

Our overall framework has three conceptual stages. The first stage is \textbf{target prediction}, whose goal is to model the intent uncertainty with a discrete set of target states $\mathcal{T}$ based on the observed context $\bfx$, and outputs the target distribution $p(\tau | \bfx)$. The second stage is \textbf{target-conditioned motion estimation}, which models the possible future motions from the initial state to the target with a unimodal distribution. The first two stages give rise to the following probabilistic predictions
$p(\bfs_F | \bfx) = \sum_{\tau \in \mathcal{T}(\bfc_P)}p(\tau | \bfx) p(\bfs_F | \tau, \bfx)$.

Many downstream applications, such as real-time behavior prediction, require a small set of representative future predictions rather than the full distribution of all possible futures. Our final stage, \textbf{scoring and selection}, is tailored for this purpose.
% where the mode of future predictions corresponding to each candidate target $\tau$ is used as the ``representative'' prediction. 
We learn a scoring function $\phi(\bfs_F)$
% which is normalized into categorical distribution
over all representative predictions, and select  a final diversified set of predictions.
\section{Target-driveN Trajectory Prediction}

\begin{figure}
    \centering
    \includegraphics[width=\linewidth]{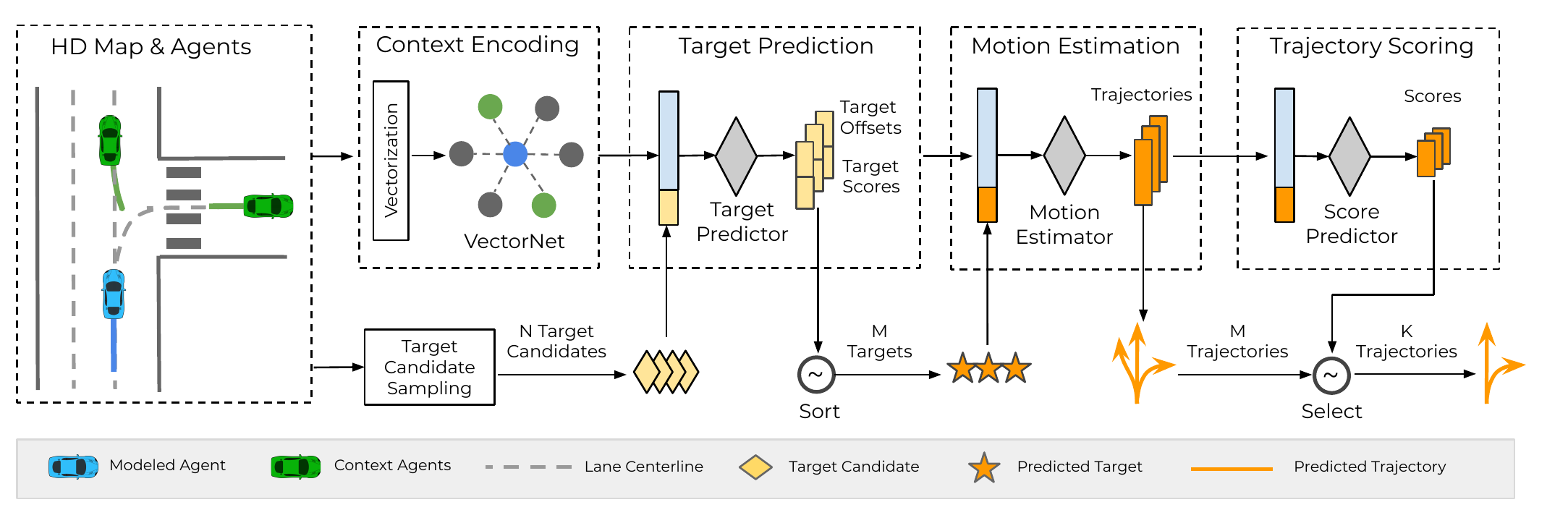}
    \caption{\model model overview. Scene context is first encoded as the model's inputs.
    %its vectorized representations.
    Then follows the core three stages of \model: (a) {\em target prediction} which proposes an initial set of $M$ targets; (b) {\em target-conditioned motion estimation} which estimates a trajectory for each target; (c) {\em scoring and selection} which ranks trajectory hypotheses and outputs a final set of $K$ predicted trajectories.}
    \label{fig:pipeline}
    % \vspace{-0.1in}
\end{figure}

This section describes our proposed \model framework in detail. We focus on the task of future trajectory prediction for moving road agents, where both the states and the targets are represented by their physical locations $(x_t, y_t)$. We begin the section by describing how the context information is encoded efficiently. We  then present details on how the proposed three stages are adapted to  the task. An overview of the \model 
model architecture is shown in Figure~\ref{fig:pipeline}.

\subsection{Scene context encoding}
\label{sec:context}

Modeling scene context is a first step in trajectory prediction so as to capture agent-road and agent-agent interactions. \model can use any suitable context encoder: when the HD map is available, we use a state-of-the-art hierarchical graph neural network VectorNet~\cite{gao2020vectornet} to encode the context. Specifically, polylines are used to abstract the HD map elements $\bfc_P$ (lanes, traffic signs) and agent trajectories $\bfs_P$; a subgraph network is applied to encode each polyline, which contains a variable number of vectors; then a global graph is used to model the interactions between polylines. 
The output is a global context feature $\bfx$ for each modeled agent.
If scene context is only available in the form of top-down imagery, a ConvNet is used as the context encoder.

\subsection{Target prediction}
\label{sec:stage1}

% Target definition
In our formulation, targets $\tau$ are defined as the locations $(x,y)$ an agent is likely to be at  a fixed time horizon $T$.
In the first \textit{target prediction} stage, we aim to provide a distribution of future targets of an agent $p(\mathcal{T}|\mathbf{x})$. We model the potential future targets via a set of $N$ discrete, quantized locations with continuous offsets: $\mathcal{T} = \{\tau^n\} = \{(x^n, y^n) + (\Delta x^n, \Delta y^n)\}_{n=1}^N$.
The distribution over targets can then be modeled via a discrete-continuous factorization:
\begin{equation}
p(\tau^n | \textbf{x}) = \pi(\tau^n | \bfx)\cdot
\mathcal{N}(\Delta x^n \;|\;\nu_x^n(\mathbf{x})) \cdot
\mathcal{N}(\Delta y^n \;|\;\nu_y^n(\mathbf{x})),
\end{equation}
where $\pi(\tau^n | \bfx) = \exp f(\tau^n, \mathbf{x}) / \sum_{\tau'} \exp f(\tau', \bfx)$ is a discrete distribution over location choices $(x^n, y^n)$. The terms $\mathcal{N}(\cdot | \nu(\cdot))$ denote a generalized normal distribution, where we choose Huber as the distance function. We denote the mean as  $\nu(\cdot)$ and assume unit variance.

The trainable functions $f(\cdot)$ and $\nu(\cdot)$ are implemented with a 2-layer multilayer perceptron (MLP), with target coordinates $(x^k, y^k)$ and the scene context feature $\bfx$ as inputs.
% (and lane feature $\bfx_l$ the target sits on if available)
They predict a discrete distribution over target locations and their most likely offsets. The loss function for training this stage is given by 
\begin{equation}
    \mathcal{L}_{\text{S1}}=\mathcal{L}_{\text{cls}}(\pi, u) + \mathcal{L}_{\text{offset}}(\nu_x, \nu_y, \Delta x^u, \Delta y^u),
\end{equation}
where $\mathcal{L}_{\text{cls}}$ is cross entropy, $\mathcal{L}_{\text{offset}}$ is the Huber loss; $u$ is the target closest to the ground truth location, and $\Delta x^u, \Delta y^u$ are the spatial offsets of $u$ from the ground truth.

\begin{figure}[t]
    \centering
    \includegraphics[width=0.65\linewidth]{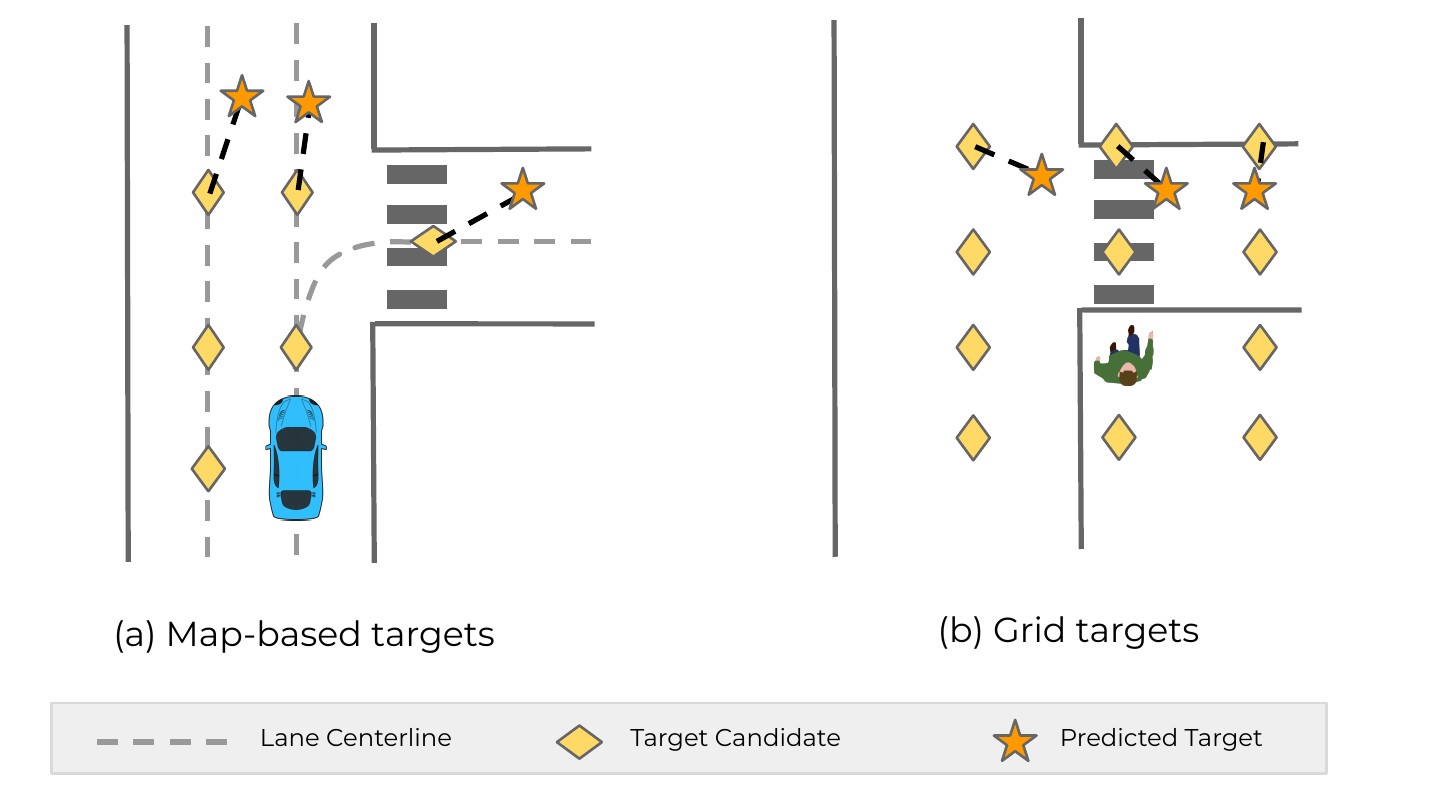}
    \caption{\model supports flexible choices of targets. Vehicle target candidate points are sampled from the lane centerlines. Pedestrian target candidate points are sampled from a virtual grid centered on the pedestrian.}
    \label{fig:targets}
\end{figure}

% Target types in specific applications
The choice of the discrete target space is flexible across different applications, as illustrated in Figure~\ref{fig:targets}. In the vehicle trajectory prediction problem, we uniformly sample points on lane centerlines from the HD map and use them as target candidates (marked as yellow spades), with the assumption that vehicles never depart far away from lanes; for pedestrians, we generate a virtual grid around the agent and use the grid points as target candidates.
For each target candidate, the \model \textit{target predictor} produces a tuple of $(\pi, \Delta x, \Delta y)$; the regressed targets are marked as orange stars.
Comparing to direct regression, the most prominent advantage of modeling the future as a discrete set of targets is that it does not suffer from mode averaging, which is the major factor that hampers multimodal predictions.

% Implementation
In practice, we over-sample a large number of target candidates as input to this stage, \eg~$N=1000$, to increase the coverage of the potential future locations; and then keep a smaller number of them as output, \eg~top $M=50$, for further processing, as a good choice of $M$ helps to balance between target recall and model efficiency.

\subsection{Target-conditioned motion estimation}
\label{sec:stage2}
In the second stage, we model the likelihood of a trajectory given a target as $p(\bfs_F | \tau, \bfx) = \prod_{t=1}^T p(s_t | \tau, \bfx)$, again with a generalized normal distribution.  This makes two assumptions. First, future time steps are conditionally independent, which makes our model computationally efficient by avoiding sequential predictions, as is done in~\cite{chai2019multipath,phan2019covernet,cui2019multimodal,mangalam2020not}.
Second, we are making strong but reasonable assumption that the distribution of the trajectories is unimodal (normal) given the target. This is certainly true for short time horizons;
for longer time horizons, one could iterate between (\textit{intermediate}) target prediction and motion estimation so that the assumption still holds.

This stage is implemented with a 2-layer MLP. It takes context feature $\bfx$ and a target location $\tau$ as input,
and outputs one most likely future trajectory $[\hat{s}_1, ..., \hat{s}_T]$ per target. Since it is conditioned on the predicted targets from the first stage, to enable a smooth learning process, we apply a teacher forcing technique~\cite{williams1989learning} at training time by feeding the ground truth location $(x^u, y^u)$ as target. The loss term for this stage is the distance between between predicted states $\mathbf{\hat{s}}_t$ and ground truth $\mathbf{s}_t$:
\begin{equation}
    \mathcal{L}_{\text{S2}}= \sum_{t=1}^T \mathcal{L}_{\text{reg}}(\mathbf{\hat{s}}_t, \mathbf{s}_t),
\end{equation}
where $\mathcal{L}_{\text{reg}}$ is implemented as Huber loss over per-step coordinate offsets.
%At inference time, the predicted $M$ targets from stage-1 are used as input to estimate $M$ trajectories.

\subsection{Trajectory scoring and selection}
\label{sec:stage3}
Our final stage estimates the likelihood of full future trajectories $\bfs_F$.  This differs from the second stage, which decomposes over time steps and targets, and from the first stage which only has knowledge of targets, but not full trajectories---\eg, a target might be estimated to have high likelihood, but a full trajectory to reach that target might not.

We use a maximum entropy model to score all the $M$ trajectories from the second stage:
\begin{equation}
    \phi(\bfs_F | \bfx) = \frac{\exp(g(\bfs_F, \bfx))}{\sum_{m=1}^M \exp(g(\bfs_F^m, \bfx))},
\end{equation}
where $g(\cdot)$ is modeled as a 2-layer MLP. 
The loss term for training this stage is the cross entropy between the predicted scores and ground truth scores,
\begin{equation}
    \mathcal{L}_{\text{S3}}= \mathcal{L}_{\text{CE}}(\phi(\bfs_F | \bfx), \psi(\bfs_F)),
\end{equation}
where the ground truth score of each predicted trajectory is defined by its distance to ground truth trajectory $\psi(\bfs_F)=\exp(-D(\bfs, \bfs_{GT}) / \alpha) / \sum_{\bfs'} \exp(-D(\bfs', \bfs_{GT}) / \alpha)$, where $D(\cdot)$ is in meters and $\alpha$ is the temperature.  The distance metric is defined as $D(\textbf{s}^i, \textbf{s}^j) = \textrm{max}(||s^i_1 - s^j_1||_2^2, ..., ||s^i_t - s^j_t||_2^2)$. 

To obtain the final small set of $K$ predicted trajectories from the scored $M$ trajectories, we implement  a trajectory selection algorithm to reject near-duplicate trajectories. We first sort the trajectories according to their score in  descending order, and then pick them greedily; if one trajectory is distant enough from all the selected trajectories, we select it as well, otherwise exclude it. The distance metric used here is the same as for the scoring process.
This process is inspired by the non-maximum suppression algorithm commonly used for computer vision problems, such as
%edge detection and
object detection.

\subsection{Training and inference details}

The above \model formulation yields fully supervised end-to-end training, with a total loss function
\begin{equation}
    \mathcal{L} = \lambda_1 \mathcal{L}_{\text{S1}} + \lambda_2 \mathcal{L}_{\text{S2}} + \lambda_3 \mathcal{L}_{\text{S3}},
\end{equation}
where $\lambda_1, \lambda_2, \lambda_3$ are chosen to balance the training process.

At inference time, \model works as follows: (1) encode context; (2) sample $N$ target candidates as input to the target predictor, take the top $M$ targets as estimated by $\pi(\tau|\bfx)$; (3) take the MAP trajectory for each of the $M$ targets from motion estimation model $p(\bfs_F | \tau, \bfx)$; (4) score the $M$ trajectories by $\phi(\bfs_F| \tau, \bfx)$, and select a final set of $K$ trajectories.

% \vspace{-0.1in}
\section{Experiments}

\subsection{Datasets}
\noindent{\bf Argoverse forecasting dataset~\cite{chang2019argoverse}} provides trajectory histories, context agents and lane centerline for future trajectory prediction. There are 333K 5-second long sequences in the dataset. The trajectories are sampled at 10Hz, with (0, 2] seconds for observation and (2, 5] seconds for future prediction. 

\noindent{\bf INTERACTION dataset~\cite{interactiondataset}} focuses on vehicle behavior prediction in highly interactive driving scenarios. It provides 4 different categories of interactive driving scenarios: roundabout (10479 vehicles), un-signalized intersection (14867 vehicles), signalized intersection (10933 vehicles), merging and lane changing (3775 vehicles).

\noindent{\bf In-house Pedestrian-at-Intersection dataset (PAID)} is an in-house pedestrian dataset collected around crosswalks and intersections. There are around 77K unique pedestrians for training and 12k unique pedestrians for test. The trajectories are sampled at 10Hz, 1-sec history trajectory is used to predict 3-sec future. Map features include crosswalks, lane boundaries and stop/yield signs.

\noindent{\bf Stanford Drone dataset (SDD)~\cite{stanford_drone_dataset}} is a video dataset with top-down recordings of college campus scenes, collected by drones.
The RGB video frames provide context similar to road maps in other datasets. We follow practice of other literature~\cite{SocialGAN,SocialLSTM,sadeghian2019sophie}, focusing on pedestrian trajectories only: frames are sampled at 2.5 Hz, 2 seconds of history (5 frames) are used as model input, and 4.8 seconds (12 frames) are the future to be predicted.

\subsection{Implementation Details}

\noindent{\bf Context encoding}. Following VectorNet~\cite{gao2020vectornet}, we convert the map elements and trajectories into a set of polylines and vectors. Each vector is represented as $[p_s, p_e, f, id_p]$, where $p_s$ and $p_e$ are start and end point of the vector, $f$ is a feature vector, which can contain feature type like lane state, and $id_p$ is the polyline index that the vector belongs to. We normalize the vector coordinates to be centered around the location of target agent at the last observed time step. After vectorization, VectorNet is used to encode context of the modeled agent, and its output feature will be consumed by \model. One exception is the Stanford Drone Dataset, which does not offer map data, we therefore use a standard ResNet-50~\cite{ResNet16} ConvNet to encode the birds-eye-view imagery for context encoding.

\noindent{\bf Target candidate sampling}. For vehicle trajectory prediction, we sample points as the target candidates from lane centerlines (Argoverse dataset) or lane boundaries (INTERACTION dataset). At least one point is sampled every meter. For pedestrian trajectory prediction, as pedestrians have much larger moving flexibility, we build a rectangular 2D grid (\eg 10m $\times$ 10m) around the agent, and the center of each cell (\eg 1m $\times$ 1m) is a target candidate.

\noindent{\bf Model details.} The model architectures of all the three stages of \model are 2-layer MLPs, with the number of hidden units set to 64. We set the temperature $\alpha$ in $\psi(\bfs_F)$ to be 0.01. The loss weights are $\lambda_1=0.1, \lambda_2=1.0, \lambda_3=0.1$. \model is trained end-to-end, for approximately 50 epochs with an Adam optimizer~\cite{kingma2014adam}. The learning rate is set to be 0.001, and batch size is 128. 

\begin{table}[h]
  \caption{Performance breakdown after each stage on the Argoverse validation set.}
  \label{tbl-breakdown}
  \centering
  \begin{tabular}{l|cccccc}
    \toprule
         &  \multicolumn{2}{c}{$\mbox{minFDE}$}  &  \multicolumn{2}{c}{$\mbox{minADE}$}  &
         \multicolumn{2}{c}{$\mbox{Miss Rate@2m}$} \\
    \midrule
        & M=50 & K=6 & M=50 & K=6 & M=50 & K=6\\
    \midrule
    S1: Target Prediction       & 0.533  & 1.629 & -    & - & 0.027 & 0.216\\
    S1 + S2: Motion Estimation   & 0.534  & 1.632 &  0.488 & 0.877 & 0.027 & 0.216 \\
    S1 + S2 + S3: Traj Scoring \& Selection   &  - & 1.292  & - & 0.728 & - & 0.093 \\
    \bottomrule
  \end{tabular}
\end{table}

\noindent{\bf Metrics.}
We adopt the widely used Average Displacement Error (ADE) and Final Displacement Error (FDE). To evaluate the ADE and FDE for a set of $K$ predicted trajectories, we use $\mbox{minADE}_K$ and $\mbox{minFDE}_K$. The displacements are all measured in meters, except for the Stanford Drone dataset where it is in pixels. On Argoverse, we also report miss rate (MR) which measures the ratio of scenarios where none of the predictions are within 2 meters of the ground truth according to FDE.

\subsection {Ablation study}

\noindent{\bf Performance breakdown by stage.}
\label{sec:perf_break_down}
We discuss the efficacy of each stage of \model by tracing the performance on the Argoverse dataset, shown in Table \ref{tbl-breakdown}.
We can see that S1 achieves good target recall as indicated by minFDE and Miss Rate at $M=50$; S2 further generates trajectories as evaluated by the minADE metric. The minFDE between S1 and S2 are almost the same, which confirms the fact that the conditional motion estimation is able to generate trajectories ending at the conditioned targets. Finally S3 narrows down the number of predictions to $K=6$ without much loss compared to $M=50$.

\noindent{\bf Target candidate sampling}.
The target candidate sampling density has an impact on TNT's performance, as shown in Table~\ref{tbl-vehicle-target} on Argoverse and Table~\ref{tbl-ped-target} on PAID respectively. For vehicles in Argoverse, we sample targets from lanes, measured as target spacing along the polyline. For pedestrians in PAID, as they have more freedom of movement, we empirically find that grid targets perform much better than map-based targets, and report only grid target results.
We observe that denser targets lead to better performance before the saturating point.

\begin{table}[h]
    \centering
    \begin{minipage}{.4\linewidth}
\caption{Comparison of map target sampling density on Argoverse dataset.}
    \label{tbl-vehicle-target}
    \centering
    \begin{tabular}{l|c|c}
    \toprule
        target spacing & $\mbox{minFDE}_{6}$ & $\mbox{minADE}_{6}$ \\ \midrule
target / 5.0m  &   1.55  &  0.79  \\ 
target / 2.0m  &   1.31  &  0.73  \\ 
target / 1.0m  &   1.29  &  0.72  \\ 
target / 0.5m &  1.29 &  0.72   \\ \bottomrule
    \end{tabular}
    \end{minipage}
    \hspace{0.5in}
    \begin{minipage}{.4\linewidth}
    \caption{Comparison of grid target sampling density on PAID.}
    \label{tbl-ped-target}
    \centering
    \begin{tabular}{l|c|c}
    \toprule
        grid size & $\mbox{minFDE}_{6}$ & $\mbox{minADE}_{6}$ \\ \midrule
        2.0m   & 0.41    &  0.22  \\ 
        1.0m &  0.33  &   0.19  \\ 
        0.5m & 0.32  &  0.18  \\ 
        0.2m &  0.32 & 0.18    \\ \bottomrule
    \end{tabular}
    \end{minipage}
\end{table}

\noindent{\bf Target regression}. The comparison between with and without target offset regression in S1 is shown in Table \ref{tbl-target}. We can see that with regression the performance improved by 0.16m, which shows the necessity of position refinement from the original target coordinates. %\vspace{-0.2in}

\begin{table}[h]
\small
    \begin{minipage}{.48\linewidth}
    \centering
    \caption{Ablation on target offset regression.}
      \label{tbl-target}
      \centering
    \begin{tabular}{l|cc}
    \toprule
            & $\mbox{minFDE}_{50}$@S1 \\ 
    \midrule
    w/ target reg  & 0.53      \\ 
    w/o target reg & 0.69     \\ 
    \bottomrule
    \end{tabular}
    \end{minipage}
    \begin{minipage}{.48\linewidth}
    \centering
    \caption{Ablation on motion estimation methods.}
      \label{tbl-ablation-s2}
      \centering
    \begin{tabular}{l|c|cc}
    \toprule
            & \# traj / target & $\mbox{minADE}_{6}$@S3  \\ 
    \midrule
    Huber & 1   &  0.73 \\ 
    % G-NLL & 1   & 0.75   \\
    CVAE & 1 & 0.73 \\
    CVAE & 10 & 0.71 \\
    \bottomrule
    \end{tabular}
    \end{minipage}
\end{table}

\begin{table}[ht]
    \caption{Comparison with state-of-the-art methods on Argoverse validation and test set. DESIRE and MultiPath are reimplemented with VectorNet context encoder. Our single model result performs on par or better than the Argoverse Challenge winner on the test set.}
      \label{tbl-argo}
      \centering
    \begin{tabular}{l|c|c c c}
    \toprule
    & subset  & $\mbox{minFDE}_{6}$ & $\mbox{minADE}_{6}$ & Miss Rate@2m\\  
    \midrule
    DESIRE~\cite{DESIRE} & \multirow{3}{*}{validation}  & 1.77 & 0.92 & 0.18\\ 
    MultiPath~\cite{chai2019multipath} & & 1.68 & 0.80 & 0.14 \\ 
    TNT (Ours) & & \textbf{1.29} & \textbf{0.73} & \textbf{0.09} \\
    \midrule
    Jean (Challenge winner) & \multirow{2}{*}{test} & \textbf{1.42} & 0.97 & \textbf{0.13} \\
    TNT (Ours) & & 1.54 & \textbf{0.94} & \textbf{0.13} \\
    \bottomrule
    \end{tabular}
\end{table}

\noindent{\bf Motion estimation methods}. For S2 motion estimation, we compare between our unimodal Huber regressor with a CVAE regressor which generates multimodal predictions. For CVAE, we vary the number of sampled trajectories between 1 and 10. The results are shown in Table~\ref{tbl-ablation-s2}. As expected, the two perform similar with only 1 trajectory. However, even when we increase the number of CVAE sampled trajectories by $10\times$, it only marginally improves on the minADE metric. This supports our assumption in S2 that the agent motion is unimodal given a target.

\begin{figure}[t]
    \centering
    \includegraphics[width=0.95\linewidth]{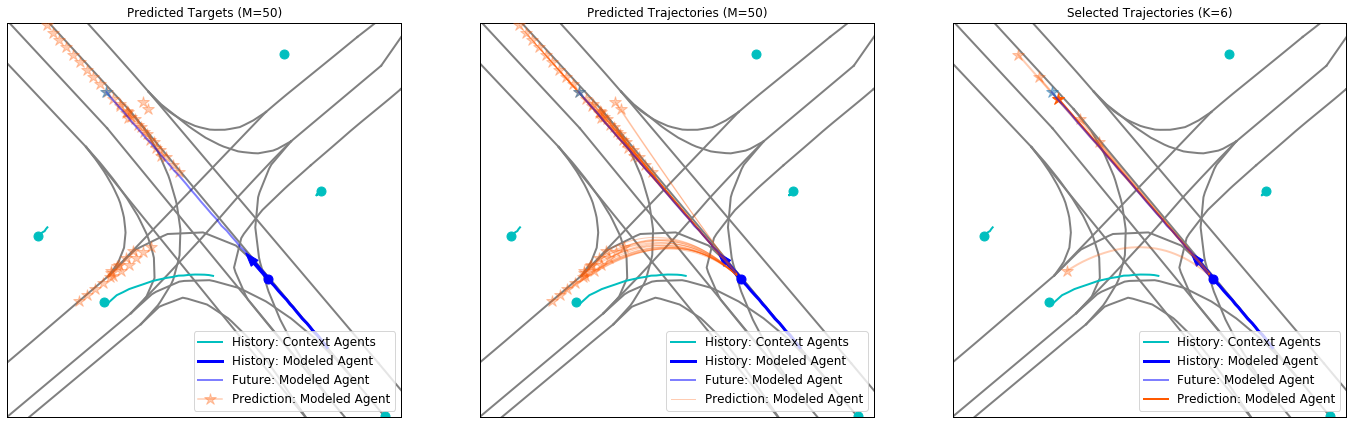}\hfill
    \includegraphics[width=0.95\linewidth]{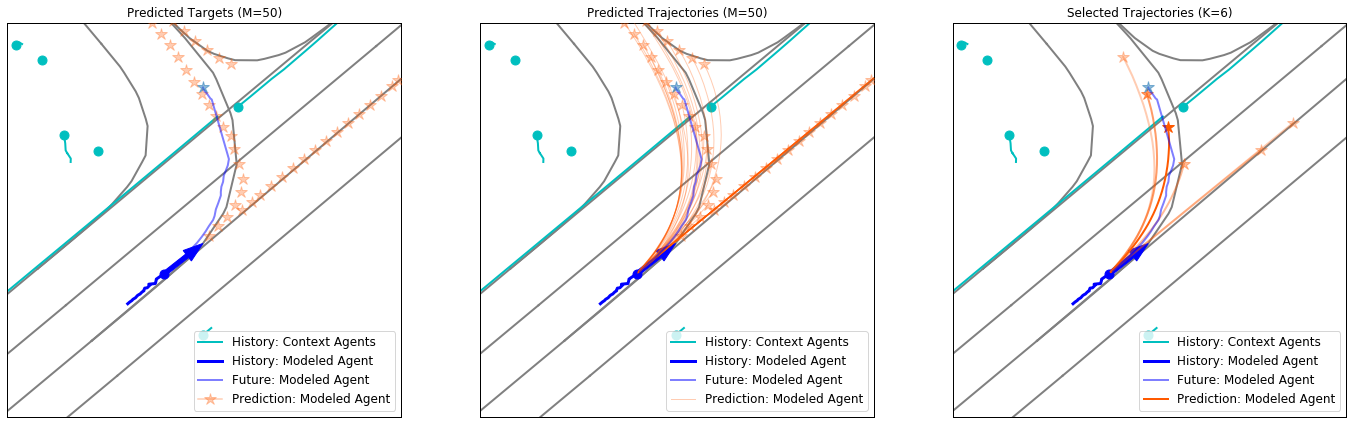}\hfill
    \includegraphics[width=0.95\linewidth]{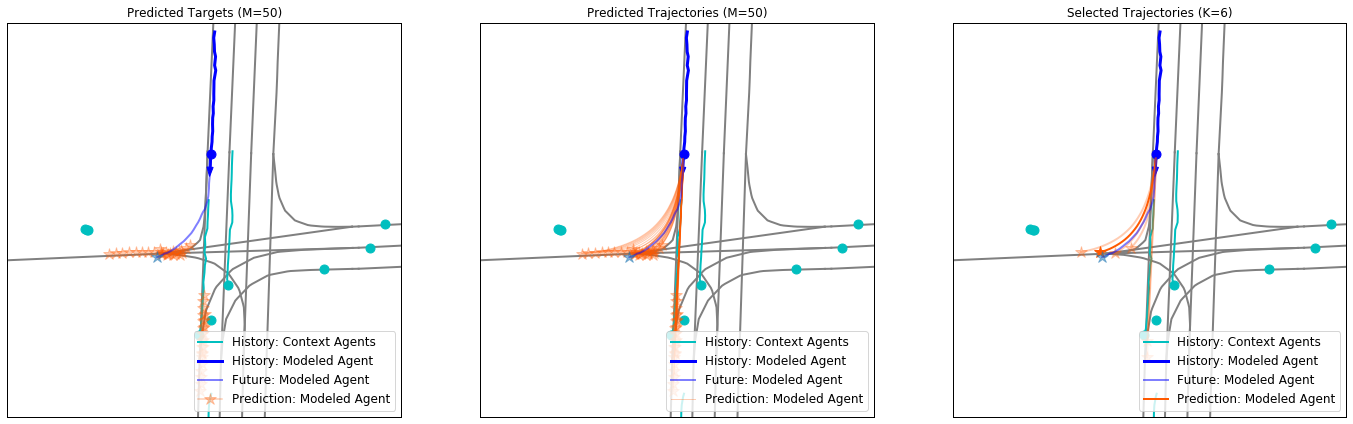}\hfill
    \includegraphics[width=0.95\linewidth]{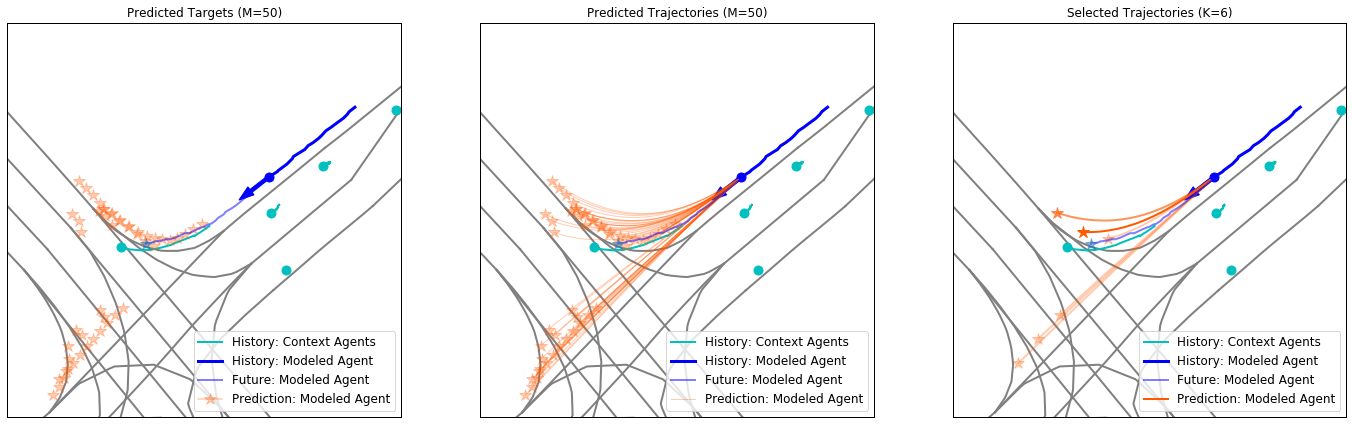}\hfill
    \includegraphics[width=0.95\linewidth]{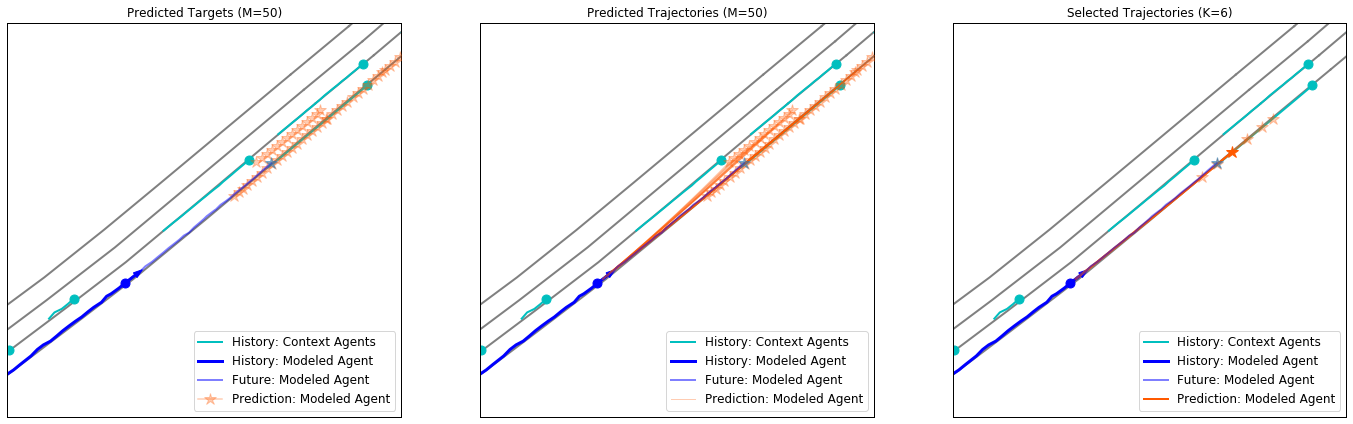}
    \caption{Qualitative results on the Argoverse validation set. Lane centerlines are shown in grey, agent's past trajectory in blue, ground truth future trajectory is in light blue. {\it(Left)} Top predicted targets, where darker color corresponds to higher scores. {\it(Middle)} Trajectory regression conditioned on the targets. {\it(Right)} Predicted trajectories after scoring and selection. The examples show \model predicting a diverse set of vehicle behaviors, among them turning, changing lanes, going straight at different speeds, \etc.}\label{fig:viz}
    %\vspace{-0.1in}
\end{figure}

\subsection{Comparison with state-of-the-art}

\noindent{\bf Vehicle prediction benchmarks}. For trajectory prediction on vehicles, we compare \model with the state-of-the-art methods on Argoverse and INTERACTION benchmarks.
To make fair comparisons, we re-implement MultiPath~\cite{chai2019multipath} and DESIRE~\cite{DESIRE} by replacing their ConvNet context encoders with VectorNet~\cite{gao2020vectornet}. As shown in Table~\ref{tbl-argo} and Table~\ref{tbl-interaction}, \model outperforms all other methods by large margins and achieves the  state-of-the-art performance. Visualizations of the \model predictions on Argoverse can be found in Figure~\ref{fig:viz}.
We further submit our single model \model results to the Argoverse leaderboard. As shown in the bottom rows in Table~\ref{tbl-argo}, %\textit{without bells and whistles},
\model performs on par or better than the Argoverse Challenge 2020 winner (its details were undisclosed).

\begin{table}[h]
    \begin{minipage}{.45\linewidth}
        \centering
        \caption{Model performance on INTERACTION validation set.}
        \label{tbl-interaction}
        \begin{tabular}{l|cc}
        \toprule
        & $\mbox{minFDE}_{6}$ & $\mbox{minADE}_{6}$ \\  \midrule
        DESIRE~\cite{DESIRE} & 0.88 & 0.32\\
        MultiPath~\cite{chai2019multipath} & 0.99 & 0.30\\
        TNT (Ours) & \textbf{0.67} & \textbf{0.21} \\\bottomrule
    \end{tabular}
    \end{minipage}
    \hspace{0.5in}
    \begin{minipage}{.45\linewidth}
        \centering
        \caption{Comparison with state-of-the-art methods on PAID.}
          \label{tbl-inhouse-stoa}
          \centering
        \begin{tabular}{l|c|c}
        \toprule
                & $\mbox{minFDE}_{3}$ & $\mbox{minADE}_{3}$ \\  \midrule
        DESIRE~\cite{DESIRE} & 0.59 & 0.29 \\
        MultiPath~\cite{chai2019multipath} & 0.43 & 0.23\\
        TNT (Ours) & \textbf{0.32} & \textbf{0.18} \\\bottomrule
        \end{tabular}
    \end{minipage} 
    %\vspace{-0.2in}
\end{table}

\noindent{\bf Pedestrian prediction benchmarks.}
For trajectory prediction on pedestrians, we compare \model with state-of-the-art methods on the in-house Pedestrian-At-Intersection Dataset (PAID) and Stanford Drone Dataset (SDD).
On the PAID, we sample targets from a grid of range $20m\times20m$ with a grid size of $0.5m$. %Similar as above, 
We enhance DESIRE and Multipath with VectorNet for context encoding.

\begin{table}
    \centering
    \caption{Comparison with state-of-the-art methods on SDD. Units are pixels.}
      \label{tbl-sdd-stoa}
      \centering
    \begin{tabular}{l|cc}
    \toprule
        & $\mbox{minFDE}_{5}$ & $\mbox{minADE}_{5}$ \\  \midrule
    Social LSTM~\cite{SocialLSTM} & 56.97 & 31.19 \\
    Social GAN~\cite{SocialGAN} & 41.44 & 27.25 \\
    DESIRE~\cite{DESIRE} & 34.05 & 19.25 \\
    SoPhie~\cite{sadeghian2019sophie} & 29.38 & 16.27 \\
    PECNet~\cite{mangalam2020not} & 25.98 & 12.79 \\
    TNT (Ours) & \textbf{21.16} & \textbf{12.23} \\\bottomrule
    \end{tabular}
\end{table}

On the SDD, since no map data is provided, we crop an image patch with resolution of $800\times800$ around the agent of interest, and use a ResNet-50 to extract context features. We use a grid of range $300\times300$ with a grid size of $6$ as targets.
As shown in Table \ref{tbl-inhouse-stoa} and Table \ref{tbl-sdd-stoa}, TNT outperforms all previous methods and achieves the state-of-the-art performance on both datasets.
\section{Conclusion}
We have presented a novel framework \model for multimodal trajectory prediction. It consists of three interpretable stages: target prediction, target-conditioned motion estimation, and trajectory scoring. \model achieves the state-of-the-art performance on four challenging real-world prediction datasets. As future work  we plan to  extend our framework to long term future prediction by iteratively predicting intermediate targets and trajectories.

\acknowledgments{We would like to thank Anca Dragan for helpful comments.}

\bibliography{ref}

\clearpage
\end{document}